# Enhancing seeding efficiency using a computer vision system to monitor furrow quality in real-time

S. Rai[1], R. Slichter[1], A. Dalal[1], and A. Sharda[1]

[1]Biological and Agricultural Engineering, Kansas State University, Manhattan, Kansas 66506, USA
sidharth@ksu.edu, asharda@ksu.edu

**Abstract**

Effective seed sowing in precision agriculture is hindered by challenges such as residue accumulation, low soil temperatures, and hair pinning (crop residue pushed in the trench by furrow opener), which obstruct optimal trench formation. Row cleaners are employed to mitigate these issues, but there is a lack of quantitative methods to assess trench cleanliness. In this study, a novel computer vision-based method was developed to evaluate row cleaner performance. Multiple air seeders were equipped with a video acquisition system to capture trench conditions after row cleaner operation, enabling an effective comparison of the performance of each row cleaner. The captured data were used to develop a segmentation model that analyzed key elements such as soil, straw, and machinery. Using the results from the segmentation model, an objective method was developed to quantify row cleaner performance. The results demonstrated the potential of this method to improve row cleaner selection and enhance seeding efficiency in precision agriculture.

**Introduction**

Machine learning has become a transformative force in precision agriculture, offering innovative tools to enhance the efficiency and accuracy of various farming processes. In recent years, machine learning techniques have been increasingly applied to optimize complex tasks in agriculture, such as crop quality control, disease detection, and weed identification (Kamath et al. 2020; Zia et al. 2022; Majeed et al. 2024). Leveraging computer vision systems for real-time monitoring and assessment of agricultural processes has unlocked new opportunities to improve seeding efficiency and optimize resource utilization.
Among these techniques, semantic segmentation—a machine learning method that enables pixel-wise classification—has proven especially valuable for identifying key components in agricultural settings, such as soil, plant material, and residue (Kamath et al. 2020). These capabilities position machine learning as an ideal solution for enhancing the seeding process, providing real-time, data-driven insights to support precision farming. One critical challenge in the seeding process is ensuring clean seed trenches, which directly impact optimal seed placement and germination.. Traditionally, the effectiveness of row cleaners has been assessed by manually collecting residue samples pre- and post-planting in drying them and weighing them. A labor intensive and time consuming task (Dadi and Raoufat 2012).
To address this limitation, this study proposes a novel approach that leverages computer vision and machine learning techniques to provide real-time, quantitative evaluations of trench cleanliness after row cleaner passage. By implementing a high-definition video data acquisition system on an air seeder row unit, detailed footage of the trenches was captured, enabling for accurate analysis of the row cleaners' effectiveness. The computer vision model developed in this study was designed to classify and quantify key elements such as soil, straw, and machinery, providing actionable insights to enhance seeding efficiency. This research not only bridges the gap in real-time trench assessment but also provides a framework for integrating automated quality evaluation tools in precision agriculture, ultimately contributing to more efficient and reliable planting outcomes. This approach offers significant potential for improving the precision of the seeding process, reducing residue interference, and ensuring optimal growing conditions for crops.

**Materials and Methods**

Data Acquisition System

The data acquisition process for the air-seeder system was carefully designed to meet the specific requirements of the study. A key component of this effort was the creation of a custom mount, illustrated in Figure 1, which was developed using 3D printing technology. Designed via CAD modeling, the mount addressed the spatial constraints of the air-seeder to ensure optimal placement. The camera's mounting angle and location were deliberately chosen to capture the entire trench scene effectively, with particular attention to the trench's cleanliness after the operation of the row cleaner. To accommodate the mount, the press wheel was removed, ensuring accurate data capture without interfering with surrounding equipment.

The mount securely housed the color camera (acA1920-40gc, Basler AG, Ahrensburg, Germany), paired with ruggedized lens (LM12HC-V, Basler AG, Ahrensburg, Germany), selected for its compatibility with the research objectives. A data acquisition system (DAQ), shown in Figure 2, was partially adapted from the system described by (Cheppally et al. 2023). The adapted setup included a fanless industrial PC (LEC-2580P-711A, Lanner Electronics, Mississauga, ON, Canada), an Ethernet-connected camera, and software from the original system, excluding certain components to better suit the study's needs. The fanless PC provided a robust and reliable platform, with adequate computational power to manage high-resolution video streams and data processing tasks. The Ethernet connection facilitated high-bandwidth data transmission, ensuring seamless integration with the software and overall system.

The Pylon Viewer software provided by Basler (*Pylon Camera Software Suite* 2023) was selected to collect the image dataset with appropriate camera settings like auto-gain and auto-exposure and shutter speed. A total of 2605 images were collected using the aforementioned system. To comprehensively evaluate the effectiveness of various row cleaners, another image dataset was collected using five different row cleaner models, including a baseline scenario with no row cleaner. A total of 5800 images were collected in this round of data collection. These image datasets were analyzed to quantify the differences in residue management effectiveness across these row cleaner configurations. This diverse dataset allowed to assess the impact of each row cleaner type on trench cleanliness. The inclusion of the baseline (without a row cleaner) provided a critical reference point, enabling a comparative analysis of how each row cleaner improves trench conditions relative to unaltered soil. This approach ensured that the developed computer vision system can not only evaluate individual row cleaners but also offer insights into the comparative advantages of different models, supporting optimized selection for precision agricultural applications.

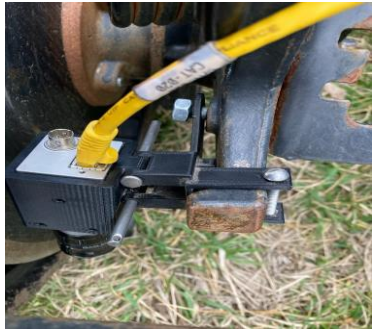 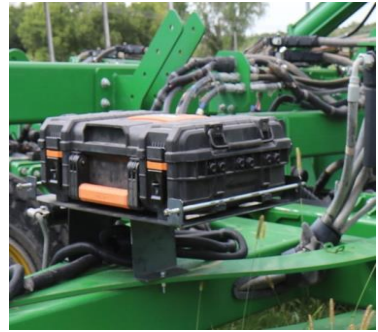

Figure 1: Air seeder mount         Figure 2: DAQ BOX

Dataset preparation

From the collected images, a total of 500 images were selected at random for the task of labeling out of 2605 images. To ensure precise labeling, RoboFlow (Dwyer et al. 2024) was utilized for the annotation process. Images were meticulously annotated to identify and classify key elements such as straw, soil, and background. The background class included any agricultural machinery visible in the frame, providing a comprehensive segmentation of the trench environment.

Labeled 500 images were split in the ratio of 80-20 for training and validation purposes, resulting in 400 images for training and 100 for validation. Image data augmentation was implemented with a goal to create a more diverse training set to ensure the model's robustness across various field conditions. Therefore, the augmentation process focused primarily on expanding the training data, increasing it to 2,000 images, while the validation set remained at 100 images ensuring that the evaluation remained unaffected by the transformations applied to the training set. This approach intentionally skewed the augmented dataset to emphasize training diversity while keeping the validation set consistent for unbiased evaluation. The augmentation generated a broader range of scenarios—by adding transformations like rotations, scaling, and color adjustments—that enriched the training set, ultimately helping the model generalize better.

Computer Vision Model Development

In this study, five different models were selected for the task of semantic segmentation: SegFormer (Xie et al. 2021), DeepLabV3+ (Chen et al. 2018), PSPNet (Zhao et al. 2017), U-Net (Ronneberger et al. 2015), and K-Net (Zhang et al. 2021). Each model was chosen for its unique strengths which are crucial for evaluating trench cleanliness in an agricultural setting. **SegFormer** features a transformer-based model that combines lightweight design with powerful global feature aggregation, making it highly efficient and robust in capturing long-range dependencies. **DeepLabV3+** leverages atrous spatial pyramid pooling (ASPP) to extract multi-scale contextual information, allowing for precise boundary delineation in cluttered environments. **PSPNet** employs pyramid pooling to capture global context, enhancing segmentation accuracy in scenarios where large-scale structures need to be identified. **U-Net** utilizes an encoder-decoder structure and skip connections, ensuring high-resolution segmentation and robustness to variations in soil texture and debris. **K-Net** introduces a kernel-based approach for dynamic instance segmentation, enabling adaptive feature extraction that is beneficial for distinguishing fine-grained details in trench images.

The training of machine learning models was carried out using the MMsegmentation framework (MMSeg.)(Contributors 2020). MMSeg is a powerful open-source toolbox based on PyTorch, was used to efficiently manage the training pipelines for all the five models to segment key elements straw, soil, and background. The models were trained using two NVIDIA RTX 4090 GPUs, allowing for parallel processing and faster model convergence. A maximum of 120 epochs was set for each model, although training was stopped early if the learning rate plateaued and became constant, indicating no further improvement in model performance. This approach ensured efficient use of computational resources and avoided overfitting. The cross entropy loss function was used as the primary loss metric Different optimizers were used for each model based on their architecture requirements: SGD was used for UNet, PSPNet and DeepLabV3+, while AdamW was employed for SegFormer and KNet. The choice of different optimizers was made to best align with the model architectures; for instance, AdamW was well-suited for transformer-based models like SegFormer due to its adaptive weight decay properties, while SGD is often effective for convolutional neural networks like PSPNet and DeepLabV3+, providing stability and better generalization for these types of architectures. The initial learning rate was adaptively reduced during training based on validation performance to ensure optimal convergence.

Quantification of Row Cleaner Performance

As the primary objective of the study was to establish a quantitative methodology for assessing row cleaner performance. The results from the developed model were used to quantify the amount of soil, straw, and background left in the trench after row cleaner passage. The percentage of each class was calculated for each frame using the following equation:

$$P^i_{class} = \frac{A^i_{class}}{A^i_{total}} * 100 \qquad (1)$$

where:
- $P^i_{class}$ : The percentage of pixels covered by the given class in frame $i$
- $A^i_{class}$: The number of pixels belonging to that class in frame $i$
- $A^i_{total}$: The total number of pixels in frame $i$

Next step was to calculate the cumulative average percentage of each class across all frames. This was done by aggregating the segmented areas for each frame and computing the cumulative average using the following equation:

$$C_{avg} = \frac{1}{N}\sum_{i=1}^{N} P^i_{class} \qquad (2)$$

where:
- $C_{avg}$: The cumulative average percentage of the given class
- $N$: The total number of frames
- $P^i_{class}$: The percentage of pixels covered by the class in frame $i$

The cumulative results were calculated to understand the overall performance of each row cleaner model in terms of soil, straw, and background removal. The results provided a quantitative assessment of each row cleaner's effectiveness in clearing the trench, enabling a robust comparison of their performance. In the end, if the percentage of soil is significantly higher than straw, it indicated that the row cleaner is doing a good job of removing straw from the trench.

**Results**

The evaluation of different computer vision models for trench cleanliness segmentation yielded notable results, as summarized in Table 1. The models were assessed on

Intersection over Union (IoU), accuracy, and inference time, providing a comprehensive understanding of their effectiveness in identifying key elements such as straw, soil, and background in the field setting. Later the best model was chosen to evaluate the trench cleanliness and row cleaner performance of multiple row cleaner models.

**Evaluation Metrics:** The evaluation of the models was conducted using three primary metrics:

- **Intersection over Union (IoU):** This metric measures the overlap between the predicted segmentation and the ground truth, providing insight into how well the model segments each class (i.e., straw, soil, background). A higher IoU indicates better model performance in correctly identifying and segmenting the classes (Everingham et al. 2010).
- **Accuracy:** Accuracy refers to the proportion of correctly classified pixels over the total number of pixels. It provides a straightforward measure of the model's overall performance in classifying the different regions (Long et al. 2015).
- **Inference Time:** This metric measures the time taken by the model to make a prediction for a given input. Lower inference time is crucial for real-time applications, where rapid decision-making is needed (Paszke et al. 2016).

Model Performance:

**SegFormer model** demonstrated the highest consistency across different classes. SegFormer, achieved notable results, with an IoU of 74.3% and accuracy of 83.07% for detecting straw, an IoU of 80.9% and accuracy of 90.89% for soil, and an IoU of 92.92% and accuracy of 97.26% for the background. The inference time was 11.75 ms, making SegFormer a balanced choice for high accuracy and relatively low latency.

**DeepLabV3+** exhibited slightly higher accuracy in detecting straw compared to SegFormer, with an IoU of 74.03% and accuracy of 87.48%. The inference time was 12.98 ms. For soil and background, DeepLabV3+ achieved an IoU of 79.65% and 92.4%, respectively, indicating consistent segmentation quality across different components.

**PSPNet** provided strong performance as well, particularly in soil and background detection. PSPNet achieved an IoU of 73.17% and accuracy of 83.99% for straw, and an IoU of 80.07% with an accuracy of 89.29% for soil. The background segmentation was also efficient, with an IoU of 92.84% and accuracy of 96.4%. The inference time for PSPNet was 20.31 ms, indicating its viability for real-time applications.

**U-Net** showed moderate performance compared to other models. It achieved an IoU of 67.69% for straw and accuracy of 83.53%, with an IoU of 72.78% and accuracy of 82.06% for soil. For the background, U-Net had an IoU of 83.3% and accuracy of 91.47%. The inference time for U-Net was 14.39 ms. While U-Net had lower segmentation metrics overall, it remains a viable choice for simpler use cases or where computational efficiency is not a priority. U-Net was also used as the baseline for segmentation performance.

**K-Net** demonstrated competitive performance, achieving an IoU of 73.91% for straw with an accuracy of 84.77%, an IoU of 80.54% for soil with an accuracy of 89.26%, and an IoU of 92.96% for the background with an accuracy of 96.87%. The inference time for K-Net was 20.05 ms, indicating its potential use in precision agricultural applications where inference speed is less critical.

Table 1. Comparison of different models to detect desired components

| Model Name | STRAW | | SOIL | | BACKGROUND | | Inference Time (ms) |
|---|---|---|---|---|---|---|---|
| | IoU | Acc | IoU | Acc | IoU | Acc | |
| SegFormer | 74.3% | 83.07% | 80.9% | 90.89% | 92.92% | 97.26% | 11.75 |
| DeepLab3 Plus | 74.03% | 87.48% | 79.65% | 86.84% | 92.4% | 96.34% | 12.98 |
| PSPNet | 73.17% | 83.99% | 80.07% | 89.29% | 92.84% | 96.4% | 20.31 |
| U-Net | 67.69% | 83.53% | 72.78% | 82.06% | 83.3% | 91.47% | 14.39 |
| K-Net | 73.91% | 84.77 | 80.54% | 89.26 | 92.96% | 96.87 | 20.05 |

Inference Time

Inference time is crucial for real-time applications in precision agriculture. **SegFormer** achieved the fastest inference time of 11.75 ms, followed closely by **DeepLabV3+** at 12.98 ms, **U-Net** at 14.39 ms, **PSPNet** at 20.31 ms, and **K-Net** at 20.05 ms. This makes SegFormer the most suitable option for integration into systems that demand near real-time feedback, with DeepLabV3+ also performing well in this regard.

Row Cleaner Evaluation

From Table 1 is clear that SegFormer is the optimal choice for real-time trench cleanliness assessment. Even though models came close the performance but SegFormer is the most well balanced model in terms of IoU, accuracy and inference time. Making it the optimal choice for real-time trench cleanliness assessment and best suited for evaluating row cleaner performance. Using the results from SegFormer, performance evaluation of different row cleaner models was done in terms of trench cleanliness using equation 1 and 2. The cumulative results are summarized in Table 2. The **No Row Cleaner** scenario showed the highest amount of straw remaining (57.55%), highlighting the significant residue left in the trench without any row cleaner intervention. **Row Cleaner C** was the most effective, leaving the least amount of straw (17.27%) and maximizing the amount of soil exposed (82.45%). **Row Cleaner B** and **Row Cleaner D** also performed well, achieving a good balance between soil exposure and straw removal.

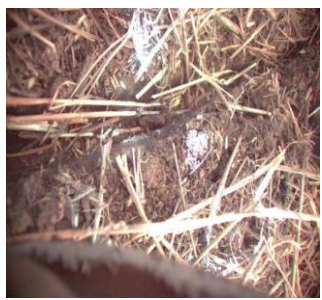 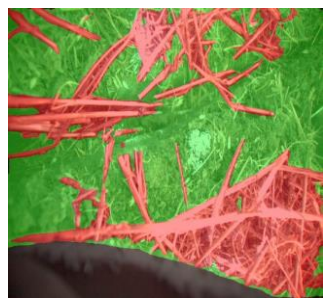 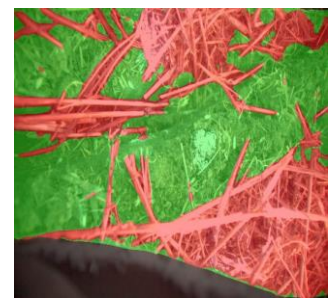

(a) Original Image.  (b) SegFormer Result.  (c) DeepLabv3+ Result

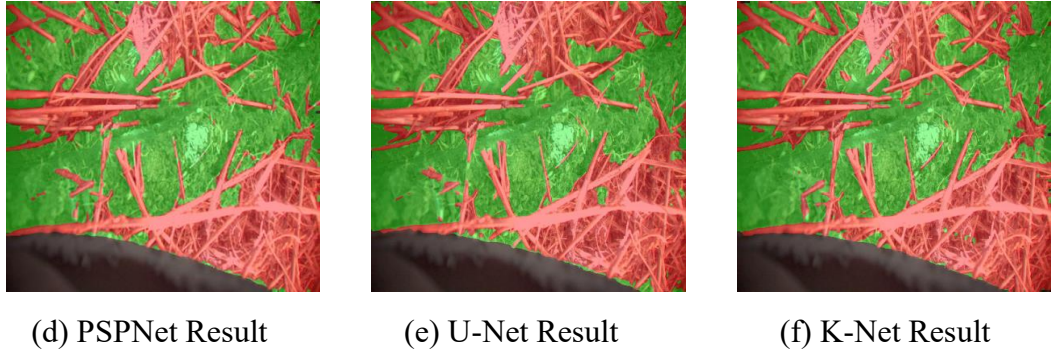

(d) PSPNet Result     (e) U-Net Result     (f) K-Net Result

Figure 3: Comparison of segmentation results across different models.

Table 2: Comparison of Row Cleaner Performance

| Row Cleaner | Amount of Soil (%) | Amount of Straw (%) | Amount of Background (%) |
|---|---|---|---|
| No Row Cleaner | 41.60 | 57.55 | 0.92 |
| Row Cleaner A | 58.35 | 40.87 | 0.84 |
| Row Cleaner B | 69.29 | 29.97 | 0.81 |
| Row Cleaner C | 82.45 | 17.27 | 0.51 |
| Row Cleaner D | 68.71 | 30.49 | 0.89 |

**Conclusion**

This study successfully developed and validated a novel, automated methodology for assessing trench cleanliness and row cleaner performance using computer vision technology. Leveraging a high-definition video acquisition system, and employing various state-of-the-art semantic segmentation models, the developed system provided a quantitative assessment of residue management efficiency across different row cleaner models. The SegFormer model emerged as the optimal choice, demonstrating superior performance in segmenting key elements—straw, soil, and background—with high accuracy and practical inference times, making it highly suitable for real-time application in agricultural settings. Through this automated approach, a consistent, objective evaluation method of trench cleanliness was offered, revealing significant differences in residue handling among row cleaner configurations. The quantitative metrics generated by this study allow for a clear and precise evaluation of row cleaner effectiveness, contributing valuable insights into optimal row cleaner selection for precision agriculture. The findings reinforce the potential of computer vision as a powerful tool for enhancing seeding operations, minimizing residue interference, and ultimately improving crop establishment. This research paves the way for more sophisticated, real-time assessments of agricultural machinery performance, thereby supporting advancements in precision agriculture and automated farming technology.